\documentclass{article}

\usepackage{microtype, graphicx, subcaption, booktabs, hyperref, tabularx, diagbox, multirow, natbib}
\usepackage{amsmath, amsfonts, amsthm}
\usepackage[all]{xy}
\usepackage[accepted]{icml2020}

\graphicspath{{./images/}}

\DeclareMathOperator{\rem}{reduce}
\DeclareMathOperator{\lcm}{lcm}
\DeclareMathOperator{\LT}{LT}
\DeclareMathOperator{\LM}{LM}
\newtheorem{thm}{Theorem}
\newtheorem{ex}{Example}
\newtheorem{defi}{Definition}
\newcommand\mybar{\kern1pt\rule[-\dp\strutbox]{.8pt}{\baselineskip}\kern1pt}

\icmltitlerunning{Learning Selection Strategies in Buchberger’s Algorithm}

\begin{document}

\twocolumn[
\icmltitle{Learning Selection Strategies in Buchberger’s Algorithm}

\begin{icmlauthorlist}
\icmlauthor{Dylan Peifer}{cornell}
\icmlauthor{Michael Stillman}{cornell}
\icmlauthor{Daniel Halpern-Leistner}{cornell}
\end{icmlauthorlist}

\icmlaffiliation{cornell}{Department of Mathematics, Cornell University, Ithaca, NY, USA}

\icmlcorrespondingauthor{Daniel Halpern-Leistner}{daniel.hl@cornell.edu}

\icmlkeywords{Reinforcement Learning, Groebner Bases, Algebraic Geometry}

\vskip 0.3in
]

\printAffiliationsAndNotice{}

\begin{abstract}
Studying the set of exact solutions of a system of polynomial equations largely depends on a single iterative algorithm, known as Buchberger’s algorithm. Optimized versions of this algorithm are crucial for many computer algebra systems (e.g., Mathematica, Maple, Sage). We introduce a new approach to Buchberger’s algorithm that uses reinforcement learning agents to perform S-pair selection, a key step in the algorithm. We then study how the difficulty of the problem depends on the choices of domain and distribution of polynomials, about which little is known. Finally, we train a policy model using proximal policy optimization (PPO) to learn S-pair selection strategies for random systems of binomial equations. In certain domains, the trained model outperforms state-of-the-art selection heuristics in total number of polynomial additions performed, which provides a proof-of-concept that recent developments in machine learning have the potential to improve performance of algorithms in symbolic computation.
\end{abstract}

\section{Introduction}\label{sec:intro}

Systems of multivariate polynomial equations, such as
\begin{equation} \label{E:system}
\left\{ \begin{array}{l} 0 = f_1(x,y) = x^3 + y^2 \\ 0 = f_2(x,y) = x^2 y-1 \end{array} \right.
\end{equation}
appear in many scientific and engineering fields, as well as many subjects in mathematics. The most fundamental question about such a system of equations is whether there exists an exact solution. If one can express the constant polynomial $h(x,y)=1$ as a combination
\begin{equation}\label{E:cancellation}
h(x,y) = a(x,y) f_1(x,y) + b(x,y) f_2(x,y)
\end{equation}
for some polynomials $a$ and $b$, then there can be no solution, because the right hand side vanishes at any solution of the system, but the left hand side is always 1.

The converse also holds: the set of solutions with $x$ and $y$ in $\mathbb C$ is empty if and only if there exists a linear combination \eqref{E:cancellation} for $h=1$ \cite{hilbert1893}. Thus the existence of solutions to \eqref{E:system} can be reduced to the larger problem of determining if a polynomial $h$ lies in the \emph{ideal} generated by these polynomials, which is defined to be the set $I = \langle f_1,f_2 \rangle$ of all polynomials of the form \eqref{E:cancellation}.

The key to solving this problem is to find a \emph{Gr\"obner basis} for the system. This is another set of polynomials $\{g_1,\ldots,g_k\}$, potentially much larger than the original set, which generate the same ideal $I = \langle f_1,f_2 \rangle = \langle g_1,\ldots,g_k \rangle$, but for which one can employ a version of the Euclidean algorithm (discussed below) to determine if $h \in I$.

In fact, computing a Gr\"{o}bner basis is the necessary first step in algorithms that answer a huge number of questions about the original system: eliminating variables, parametrizing solutions, studying geometric features of the solution set, etc. This has led to a wide array of scientific applications of Gr\"obner bases, wherever polynomial systems appear, including: computer vision \cite{DuffEtAl2019}, cryptography \cite{AppCryptology2010}, biological networks and chemical reaction networks \cite{Arkun2019},  robotics \cite{Ablamowicz2010}, statistics \cite{DiaconisSturmfels1998, Sullivant2018}, string theory \cite{Gray2011}, signal and image processing \cite{LinXuWu04}, integer programming \cite{Conti1991}, coding theory \cite{SalaEtAl2009}, and splines \cite{CoxEtAl2005}.

Buchberger's algorithm \cite{Buchberger1965a, Buchberger1965b} is the basic iterative algorithm used to find a Gr\"obner basis. As it can be costly in both time and space, this algorithm is the computational bottleneck in many applications of Gr\"obner bases.
All direct algorithms for finding Gr\"obner bases (e.g., \cite{FaugereF4,FaugereF5,RouneStillman2012,EderFaugere2017}) are variations of Buchberger's algorithm, and highly optimized versions of the algorithm are a key piece of computer algebra systems such as \cite{CoCoA,Macaulay2,Magma,Maple,Mathematica,Sagemath,Singular}.

There are several points in Buchberger's algorithm which depend on choices that do not affect the correctness of the algorithm, but can have a significant impact on performance. In this paper we focus on one such choice, called \emph{pair selection}.
We show that the problem of pair selection fits naturally into the framework of reinforcement learning, and claim that the rapid advancement in applications of deep reinforcement learning over the past decade has the potential to significantly improve the performance of the algorithm.

Our main contributions are the following:
\begin{enumerate}
    \item Initiating the empirical study of Buchberger's algorithm from the perspective of machine learning.
    \item Identifying a precise sub-domain of the problem, consisting of systems of binomials, that is directly relevant to applications, captures many of the challenging features of the problem, and can serve as a useful benchmark for future research.
    \item Training a simple neural network model for pair selection which outperforms state-of-the art selection strategies by $20\%$ to $40\%$ in this domain, thereby demonstrating significant potential for future work.
\end{enumerate}

\subsection{Related Work}

Several authors have applied machine learning to perform algorithm selection \cite{HuangEtAl2019} or parameter selection \cite{Xu2019OnlineSI} in problems related to Gr\"{o}bner bases. While we are not aware of any existing work applying machine learning to improve the performance of Buchberger's algorithm, many authors have used machine learning to improve algorithm performance in other domains \cite{AlvarezEtAl2017,KhalilEtAl2016}. Recently, there has been progress using reinforcement learning to learn entirely new heuristics and strategies inside algorithms \cite{BengioEtAl2018}, which is closest to our approach.


\section{Gr\"{o}bner Bases}\label{sec:groebnerbases}

In this section we give a focused introduction to Gr\"{o}bner basis concepts that will be needed for Section~\ref{sec:buchbergerenv}.
For a more general introduction to Gr\"{o}bner bases and their uses, see \cite{CoxEtAl2015,Mora2005}.

Let $R = K[x_1, \dots, x_n]$ be the set of polynomials in variables $x_1, \dots, x_n$ with coefficients in some field $K$.
Let $F = \{f_1, \dots, f_s\}$ be a set of polynomials in $R$, and consider $I = \langle f_1, \dots, f_s \rangle$ the ideal generated by $F$ in $R$.

The definition of a Gr\"obner basis depends on a choice of \emph{monomial order}, a well-order relation $>$ on the set of monomials $\{x^a = x_1^{a_1}\cdots x_n^{a_n}| a \in \mathbb{Z}^n_{\geq 0} \}$ such that $x^a > x^b$ implies $x^{a+c} > x^{b+c}$ for any exponent vectors $a,b,c$. Given a polynomial $f = \sum_{a} \lambda_a x^a$, we define the \emph{leading term} $\LT(f) = \lambda_a x^a$, where the \emph{leading monomial} $\LM(f) = x^a$ is the largest monomial with respect to the ordering $>$ that has $\lambda_a \neq 0$.  An important example is the \emph{grevlex} order, where $x^a > x^b$ if the total degree of $x^a$ is greater than that of $x^b$, or they have the same degree, but the \emph{last} non-zero entry of $a-b$ is \emph{negative}. For example, in the grevlex order, we have $x_1 > x_2 > x_3$, $x_2^3 > x_1 x_2$, and $x_2^2 > x_1 x_3$.

Given a choice of monomial order $>$ and a set of polynomials $F = \{f_1,\ldots, f_s\}$, the \emph{multivariate division algorithm} takes any polynomial $h$ and produces a remainder polynomial $r$, written $r = \rem(h, F)$, such that $h-r \in \langle f_1,\ldots,f_s \rangle$ and $\LT(f_i)$ does not divide $\LT(r)$ for any $i$. In this case we say that $h$ \emph{reduces to }$r$.
The division algorithm is guaranteed to terminate, but the remainder can depend on the choice in line 5 of Algorithm~\ref{alg:division}.

\begin{algorithm}[tb]
   \caption{Multivariate Division Algorithm}
   \label{alg:division}
\begin{algorithmic}[1]
   \STATE {\bfseries Input:} a polynomial $h$ and a set of polynomials $F = \{f_1, \dots, f_s\}$
   \STATE {\bfseries Output:} a remainder polynomial $r = \rem(h, F)$
   \STATE $r \gets h$
   \WHILE{$\LT(f_i)|\LT(r)$ for some $i$}
      \STATE choose $i$ such that $\LT(f_i)|\LT(r)$
      \STATE $r \gets r - \frac{\LT(r)}{\LT(f_i)} f_i$
    \ENDWHILE
\end{algorithmic}
\end{algorithm}

\begin{defi}\label{defi:gb}
Given a monomial order, a \emph{Gr\"{o}bner basis} $G$ of a nonzero ideal $I$ is a set of generators $\{g_1, g_2, \dots, g_k\}$ of $I$ such that any of the following equivalent conditions hold:
\begin{align*}
\text{(i)}\quad & \rem(h, G) = 0 \iff h \in I \\
\text{(ii)}\quad & \rem(h, G) \text{ is unique for all $h \in R$} \\
\text{(iii)}\quad & \langle \LT(g_1), \LT(g_2), \dots, \LT(g_k) \rangle = \langle \LT(I) \rangle
\end{align*}
where $\langle\LT(I)\rangle = \langle\LT(f) \,|\, f \in I\rangle$ is the ideal generated by the leading terms of all polynomials in $I$.
\end{defi}

As mentioned in Section~\ref{sec:intro}, a consequence of $(i)$ is that given a Gr\"obner basis $G$ for $\langle f_1,\ldots, f_s\rangle$, the system of equations $f_1 = 0, \dots, f_s = 0$ has no solution over $\mathbb{C}$ if and only if $\rem(1, G) = 0$, that is, if $G$ contains a non-zero constant polynomial.

\subsection{Buchberger's Algorithm}

Buchberger's algorithm produces a Gr\"{o}bner basis for the ideal $I = \langle f_1, \dots, f_s \rangle$ from the initial set $\{f_1, \dots, f_s\}$ by repeatedly producing and reducing combinations of the basis elements.

\begin{defi}
Let $S(f, g) = \frac{x^\gamma}{\LT(f)} f - \frac{x^\gamma}{\LT(g)} g$, where $x^\gamma = \lcm(\LM(f), \LM(g))$ is the least common multiple of the leading monomials of $f$ and $g$.
This is the \emph{$S$-polynomial} of $f$ and $g$, where $S$ stands for subtraction or syzygy.
\end{defi}

\begin{thm}[Buchberger's Criterion]\label{thm:buchberger}
Suppose the set of polynomials $G = \{g_1, g_2, \dots, g_k\}$ generates the ideal $I$.
If $\rem(S(g_i, g_j), G) = 0$ for all pairs $g_i, g_j$ then $G$ is a Gr\"{o}bner basis of $I$.
\end{thm}

\begin{algorithm}[b]
   \caption{Buchberger's Algorithm}
   \label{alg:buchberger}
\begin{algorithmic}[1]
   \STATE {\bfseries Input:} a set of polynomials $\{f_1, \dots, f_s\}$
   \STATE {\bfseries Output:} a Gr\"{o}bner basis $G$ of $I = \langle f_1, \dots, f_s \rangle$
   \STATE $G \gets \{f_1, \dots, f_s\}$
   \STATE $P \gets \{(f_i, f_j) : 1 \le i < j \le s\}$
   \WHILE{$|P| > 0$}
      \STATE $(f_i, f_j) \gets \mathrm{select}(P)$
      \STATE $P \gets P \setminus \{(f_i, f_j)\}$
      \STATE $r \gets \rem(S(f_i, f_j), G)$
      \IF {$r \neq 0$}
        \STATE $P \gets \mathrm{update}(P, G, r)$
        \STATE $G \gets G \cup \{r\}$
      \ENDIF
    \ENDWHILE
\end{algorithmic}
\end{algorithm}

\begin{ex}
Fix $>$ to be grevlex.
For the generating set $F = \{f_1,f_2\}$ in Equation \eqref{E:system}, $r = \rem(S(f_1,f_2), F) = y^3+x$.
By construction, the set $G = \{f_1,f_2,r\}$ generates the same ideal as $F$ and $\rem(S(f_1,f_2), G) = 0$, so we have eliminated this pair for the purposes of verifying the criterion at the expense of two new pairs.
Luckily, in this example $\rem(S(f_1,r), G) = 0$ and $\rem(S(f_2,r), G) = 0$, so $G$ is a Gr\"obner basis for $\langle f_1, f_2\rangle$ with respect to the grevlex order.
\end{ex}

Generalizing this example, Theorem~\ref{thm:buchberger} naturally leads to Algorithm~\ref{alg:buchberger}, which depends on several implementation choices: {\tt select} in line 6, {\tt reduce} in line 8, and {\tt update} in line 10.
Algorithm~\ref{alg:buchberger} is guaranteed to terminate regardless of these choices, but all three impact computational performance.
Most improvements to Buchberger's algorithm have come from improved heuristics in these steps.

The simplest implementation of {\tt update} is
\[ \mathrm{update}(P, G, r) = P \cup \{(f, r) : f \in G\}, \]
but most implementations use special rules to eliminate some pairs a priori, so as to minimize the number of $S$-polynomial reductions performed.
In fact, much recent research on improving the performance of Buchberger's algorithm \cite{FaugereF5, EderFaugere2017} has focused on mathematical methods to eliminate as many pairs as possible.
We use the standard pair elimination rules of \cite{GebauerMoeller1988} in all results in this paper.

The main choice in {\tt reduce} occurs in line 5 of Algorithm~\ref{alg:division}.
For our experiments, we always choose the smallest $\LT(f_i)$ which divides $r$.
We also modify Algorithm~\ref{alg:division} to fully \emph{tail reduce}, which leaves no term of $r$ divisible by any $\LT(f_i)$.

Our focus is the implementation of {\tt select}.

\subsection{Selection Strategies}

The selection strategy, which chooses the pair $(f_i, f_j)$ to process next, is critically important for efficiency, as poor pair selection can add many unnecessary elements to the generating set before finding a Gr\"obner basis.
While there is some research on selection \cite{FaugereF5,RouneStillman2012}, most is in the context of signature Gr\"obner bases and Faugere's $F_5$ algorithm.
Other than these, most strategies to date depend on relatively simple human-designed heuristics.
We use several well-known examples as benchmarks:

\textbf{First}: Among pairs with minimal $j$, select the one with minimal $i$.
In other words, treat the pair set $P$ as a queue.

\textbf{Degree}: Select the pair with minimal total degree of $\lcm(\LM(f_i), \LM(f_j))$.
If needed, break ties with First.

\textbf{Normal}: Select the pair with $\lcm(\LM(f_i), \LM(f_j))$ minmal in the monomial order.
If needed, break ties with First. In a degree order ($x^a > x^b$ if the total degree of $x^a$ is greater than that of $x^b$), this is a refinement of Degree selection.

\textbf{Sugar}: Select the pair with minimal sugar degree, which is the degree $\lcm(\LM(f_i), \LM(f_j))$ would have had  if all input polynomials were homogenized. If needed, break ties with Normal. Presented in \cite{GioviniEtAl1991}.

\textbf{Random}: Select an element of the pair set uniformly at random.

Most implementations use Normal or Sugar selection.

\subsection{Complexity} \label{sect:complexity}


We will characterize the input to Buchberger's algorithm in terms of the number of variables ($n$), the maximal degree of a generator ($d$), and the number of generators ($s$).
One measure of complexity is the maximum degree $\deg_{\max}(GB(I))$ of an element in the \emph{unique} reduced minimal Gr\"obner basis for $I$.


When the coefficient field has characteristic $0$, there is an upper bound $\deg_{\max}(GB(I)) \le (2d)^{2^{n+1}}$ which is double exponential in the number of variables \cite{BayerMumford93}. There do exist ideals which exhibit double exponential behavior
 \cite{MayrMeyer82, BayerStillman88, Koh98}: there
is a sequence of ideals $\{J_n\}$ where $J_n$ is generated by quadratic homogeneous binomials in $22n-1$ variables such that for any monomial order
\[ 2^{2^{n-1}-1} \le  \deg_{\max}(GB(J_n)) \]

In the grevlex monomial order, the theoretical upper bounds on the complexity of Buchberger's algorithm are much better if the choice of generators is sufficiently generic. To make this precise, for fixed $n,d,s$, the space of possible inputs, i.e., the space $V$ of coefficients for each of the $s$ generators, is finite dimensional. There is a subset $X \subset V$ of measure zero\footnote{Technically, $X$ is a closed algebraic subset. With coefficients in $\mathbb{R}$ or $\mathbb{C}$, this is measure zero in the usual sense.} such that for any point outside $X$,
\[ \deg_{\max}(GB(I)) \le (n+1)(d-1) + 1 \]
This implies that the size of $GB(I)$ is less than or equal to the number of monomials of degree less than or equal to $(n+1)(d-1) + 1$, which grows like $\mathcal{O}(((n+1)d-1)^n)$.


It is expected, but not known, that it is rare for the maximum degree of a Gr\"obner basis element in the grevlex monomial order to be double exponential in the number of variables. Also, as early as the 1980's, it was realized that for many examples, the grevlex Gr\"obner basis was often much easier to compute than Gr\"obner bases for other monomial orders.  For these reasons, the grevlex monomial order is a standard choice in Gr\"obner basis computations. We use grevlex throughout this paper for all of our experiments.

\section{The Reinforcement Learning Problem}\label{sec:buchbergerenv}

We model Buchberger's algorithm as a Markov Decision Process (MDP) in which an agent interacts with an environment to perform pair selection in line 6 of Algorithm~\ref{alg:buchberger}.

Each pass through the {\tt while} loop in line 5 of Algorithm~\ref{alg:buchberger} is a time step, in which the agent takes an action and receives a reward. At time step $t$, the agent's state $s_t = (G_t,P_t)$ consists of the current generating set $G_t$ and the current pair set $P_t$. The agent must select a pair from the current set, so the set of allowable actions is $\mathcal{A}_t = P_t$. Once the agent selects an action $a_t \in \mathcal{A}_t$, the environment updates by removing the pair from the pair set, reducing the corresponding $S$-polynomial, and updating the generator and pair set if necessary.

After the environment updates, the agent receives a reward $r_t$ which is $-1$ times the number of polynomial additions performed in the reduction of pair $a_t$, including the subtraction that produced the $S$-polynomial. This is a proxy for computational cost that is implementation independent, and thus useful for benchmarking against other selection heuristics. For simplicity, this proxy does not penalize monomial division tests or computing pair eliminations.


Each trajectory $\tau = (s_0, a_0, r_1, s_1, \dots, r_T)$ is a sequence of steps in Buchberger's algorithm, and ends when the pair set is empty and the algorithm has terminated with a Gr\"{o}bner basis. The agent's objective is to maximize the expected return ${\mathbb E}[\sum_{t=1}^T \gamma^{t-1} r_t]$, where $0 \le \gamma \le 1$ is a discount factor. With $\gamma=1$, this is equivalent to minimizing the expected number of polynomial additions taken to produce a Gr\"{o}bner basis.

This problem poses several interesting challenges from a machine learning perspective:
\begin{enumerate}
    \item The size of the action set changes with each time step and can be very large.
    \item There is a high variance in difficulty of problems of the same size.
    \item The state changes shape with each time step, and the state space is unbounded in several dimensions: number of variables, degree and size of generators, number of generators, and size of coefficients.
\end{enumerate}

\subsection{The Domain: Random Binomial Ideals}

Formulating Buchberger's algorithm as a reinforcement learning problem forces one to consider the question of what is a random polynomial. This is a significant departure from the typical framing of the Gr\"obner basis problem.

We have seen that Buchberger's algorithm performs much better than its worst case on generic choices of input. On the other hand, many of the ideals that arise in practice are far from generic in this sense. As $n,d,$ and $s$ grow, Gr\"obner basis computations tend to blow up in several ways simultaneously: (i) the number of polynomials in the generating set grows, (ii) the number of terms in each polynomial grows, and (iii) the size of the coefficients grows (e.g., rational numbers with very large denominators).

The standard way to handle (iii) in evaluating Gr\"{o}bner basis algorithms is to work over the finite field $\mathbb{F}_p = \mathbb{Z}/p\mathbb{Z}$ for a large prime number $p$. The choice $\mathbb{Z}/32003\mathbb{Z}$ is common, if seemingly arbitrary, and all of our experiments use this coefficient field. Finite field coefficients are already of use in many applications \cite{bettale13}. They also figure prominently in many state of the art Buchberger implementations with rational coefficients: the idea is to start with a generating set with integer coefficients, reduce mod $p$ for several large primes, compute the Gr\"{o}bner bases for each of the resulting systems over finite fields, then ``lift" these Gr\"{o}bner bases back to rational polynomials \cite{Arnold2003}.

In order to address (ii), we restrict our training to systems of polynomials with at most two terms. These are known as \emph{binomials}. We will also assume neither term is a constant. If the input to Buchberger's algorithm is a set of binomials of this form, then all of the new generators added to the set will also have this form. This side-steps the thorny issue of how to represent a polynomial of arbitrary size to a neural network.

Restricting our focus to binomial ideals has several other benefits: We will show that using binomials typically avoids the known ``easy" case when the dimension of the ideal, which is defined to be the dimension of the set of solutions of the corresponding system of equations, is zero. We have also seen that some of the worst known examples with double exponential behavior are binomial systems. Finally, binomials capture the qualitative fact that many of the polynomials appearing in applications are sparse. In fact, several applications of Buchberger's algorithm, such as integer programming, specifically call for binomial ideals \cite{CoxEtAl2005,Conti1991}.

We also remark that a model trained on binomials might be useful in other domains as well. Just as most standard selection strategies only consider the leading monomials of each pair, one could use a model trained on binomials to select pairs based on their leading \emph{binomials}.

We performed experiments with two probability distributions on the set of binomials of degree $\leq d$ in $s$ generators. The first, {\tt weighted}, selects the degree of each monomial uniformly at random, then selects each uniformly at random among monomials of the chosen degree. The second, {\tt uniform}, selects both monomials uniformly at random from the set of monomials of degree $\leq d$. The main difference between these two distributions is that {\tt weighted} tends to produce more binomials of low total degree. Both distributions assign non-zero coefficients uniformly at random.

For the remainder of the paper, we will use the format ``$n$-$d$-$s$ (uniform/weighted)" to specify our distribution on $s$-tuples of binomials of degree $\leq d$ in $n$ variables.

\subsection{Statistics}

We will briefly discuss the statistical properties of the problem in the domain of binomial ideals to highlight its features and challenges.

\textbf{Difficulty increases with $n$:} (Table \ref{tab:selectionstrategies}) This is consistent with the double exponential behavior in the worst-case analysis.

\textbf{Degree and Normal outperform First and Sugar:} (Table \ref{tab:selectionstrategies}) This pattern is consistent across all distributions in the range tested ($n=3$, $d\leq30$, $s\leq 20$). The fact that Sugar under-performs in an average-case analysis might reflect the fact that it was chosen because it improves performance on known sequences of challenging benchmark ideals in \cite{GioviniEtAl1991}.

\begin{figure}[h]
  \vskip 0.2in
  \begin{center}
  \centerline{\includegraphics[width=\columnwidth]{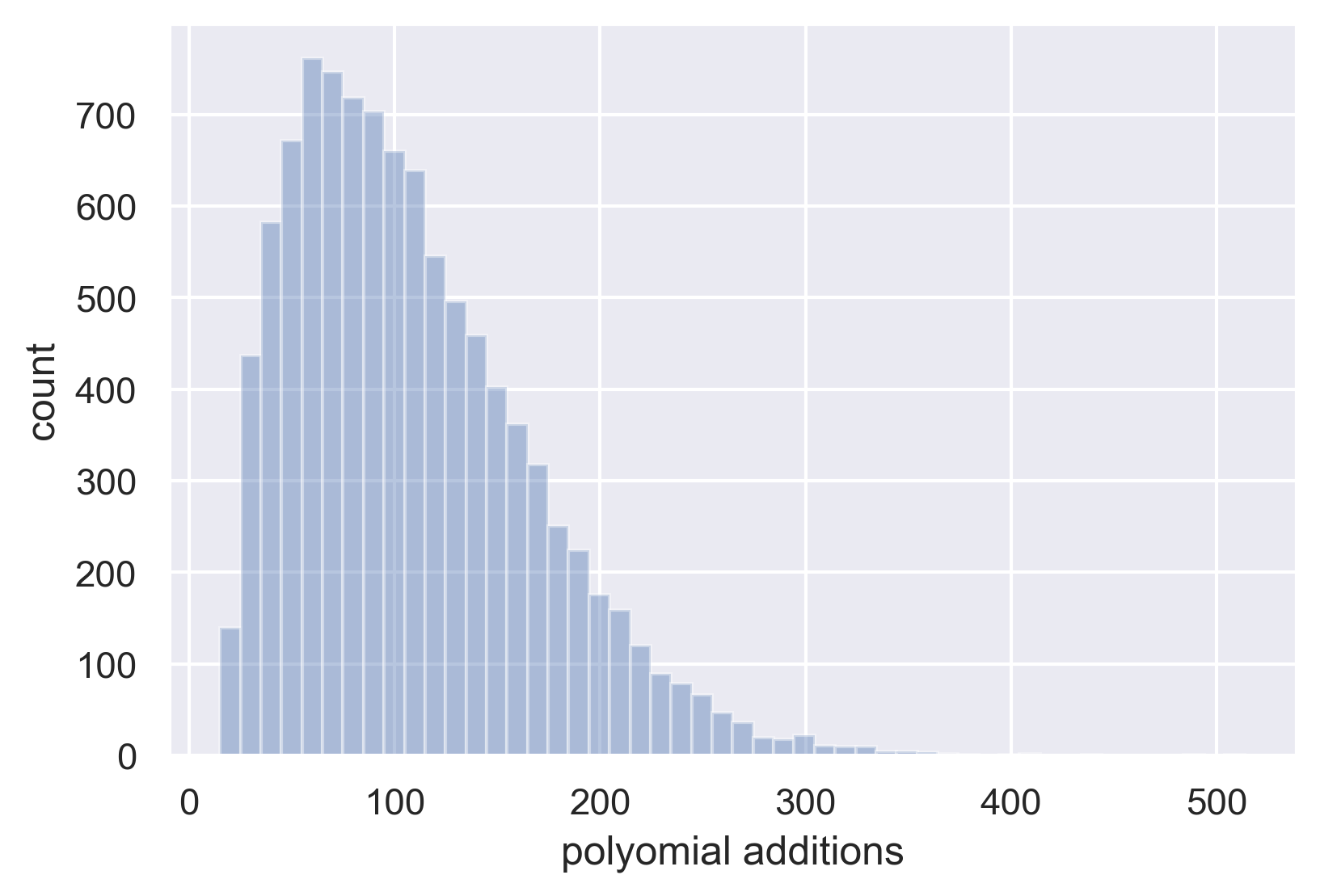}}
  \caption{Histogram of polynomial additions in 5-5-10 weighted following Degree selection over 10000 samples.}
  \label{fig:hist}
  \end{center}
  \vskip -0.2in
\end{figure}

\textbf{Very high variance in difficulty:} This is also illustrated in Table \ref{tab:selectionstrategies}, especially as the number of variables increases. Figure \ref{fig:hist} provides a more detailed view of a single distribution, demonstrating the large variance and long right tail that is typical of Gr\"obner basis calculations. This poses a particular challenge for the training of reinforcement learning models.

\begin{table}[h]
  \caption{Number of polynomial additions for different selection strategies on the same samples of 10000 ideals. Distributions are $n$-5-10 weighted. Table entries show mean [stddev].}
  \label{tab:selectionstrategies}
  \vskip 0.15in
  \begin{center}
  \begin{small}
  \begin{sc}
  \begingroup
  \setlength{\tabcolsep}{5pt}
  \begin{tabular}{r|cccc}
  \toprule
  $n$ & First & Degree & Normal & Sugar \\
  \midrule
  2 & 36.4 [7.24] & 32.3 [5.71] & 32.0 [5.49] & 32.4 [6.15]\\
  3 & 52.8 [17.9] & 42.2 [13.2] & 42.4 [13.1] & 44.2 [15.1]\\
  4 & 86.3 [40.9] & 63.8 [28.5] & 66.5 [29.8] & 70.0 [32.9]\\
  5 & 151. [85.7] & 109. [58.8] & 117. [64.4] & 120. [68.7]\\
  6 & 280. [174.] & 198. [118.] & 221. [132.] & 223. [143.]\\
  7 & 527. [359.] & 379. [240.] & 435. [277.] & 430. [296.]\\
  8 & 1030 [759.] & 760. [510.] & 887. [588.] & 863. [639.]\\
  \bottomrule
  \end{tabular}
  \endgroup
  \end{sc}
  \end{small}
  \end{center}
  \vskip -0.1in
\end{table}

\begin{figure}[ht]
  \vskip 0.2in
  \begin{center}
  \centerline{\includegraphics[width=\columnwidth]{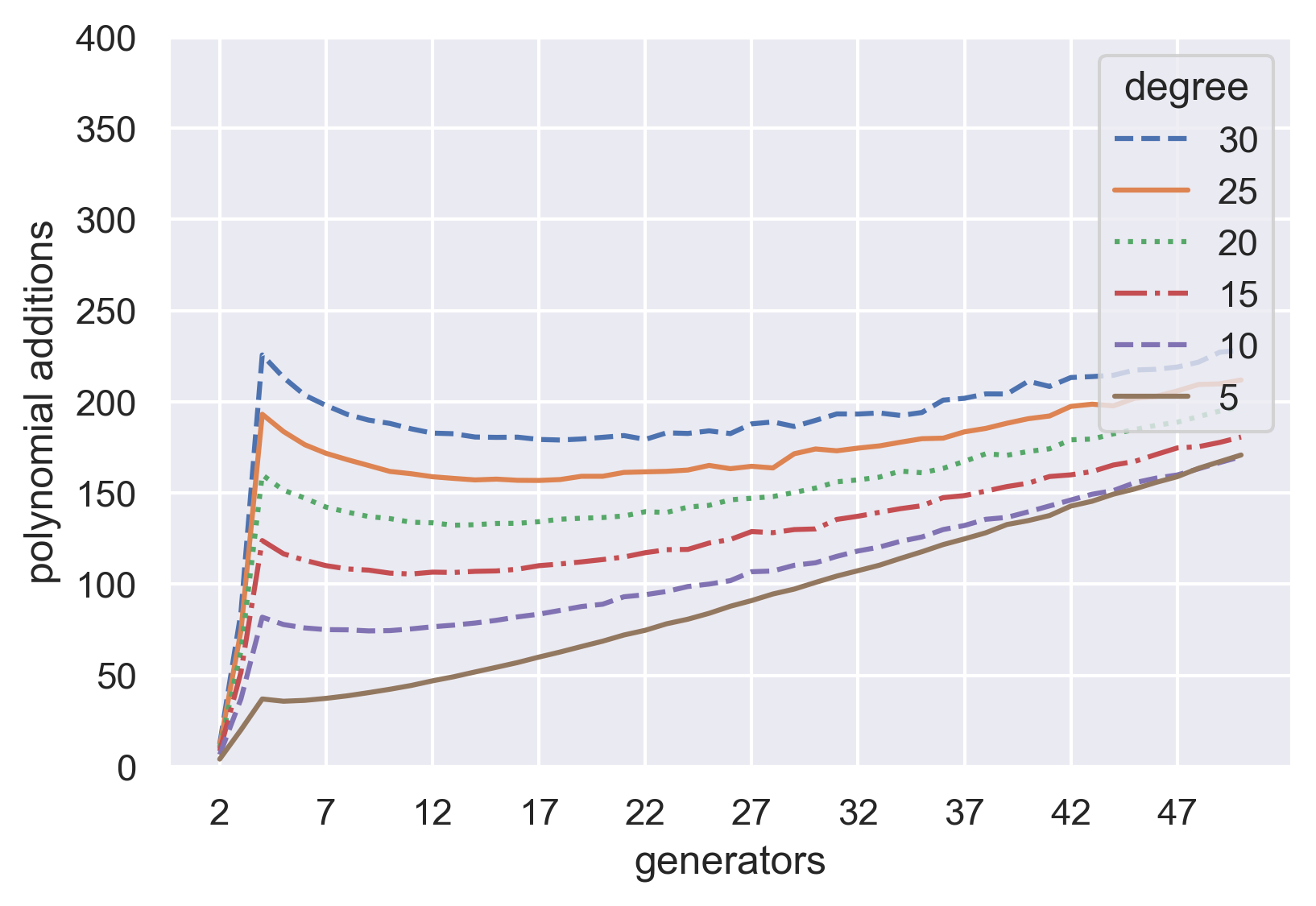}}
  \caption{Average number of polynomial additions following Degree selection in $n=3$ weighted. Each degree and generator point is the mean over 10000 samples for $s \le 20$ and 1000 samples for $s > 20$.}
  \label{fig:3-weighted-additions}
  \end{center}
  \vskip -0.2in
\end{figure}

\begin{figure}[ht]
  \vskip 0.2in
  \begin{center}
  \centerline{\includegraphics[width=\columnwidth]{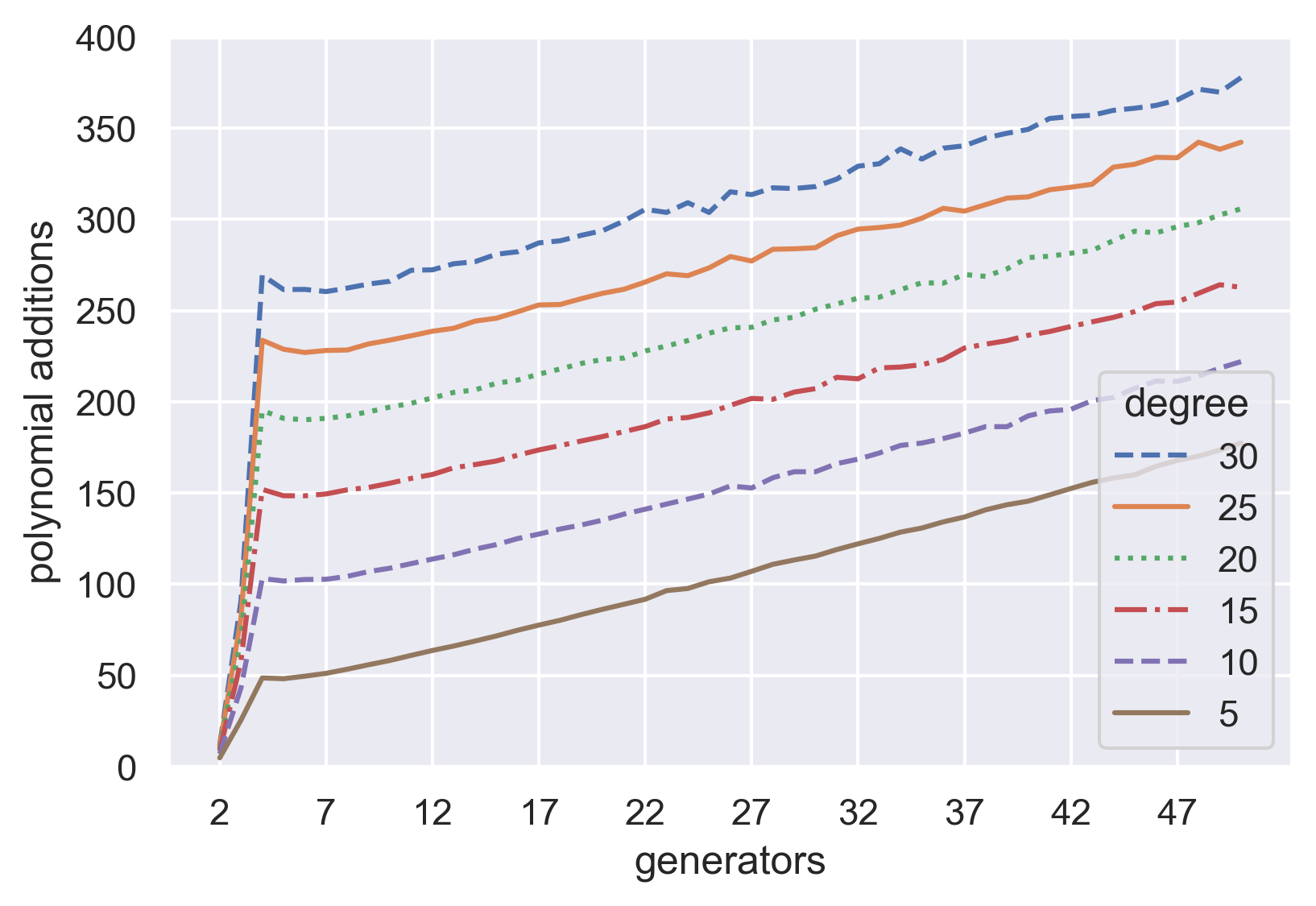}}
  \caption{Average number of polynomial additions following Degree selection in $n = 3$ uniform. Each degree and generator point is the mean over 10000 samples for $s \le 20$ and 1000 samples for $s > 20$.}
  \label{fig:3-uniform-additions}
  \end{center}
  \vskip -0.2in
\end{figure}

\textbf{Dependence on $s$ is subtle:} For $n=3$, there is is a spike in difficulty at four generators, followed by a drop/leveling off, and a slow increase after that (Figures \ref{fig:3-weighted-additions} and \ref{fig:3-uniform-additions}). The spike is even more pronounced in $n>3$ variables, where it occurs instead at $n+1$ generators. The leveling off is consistent with the hypothesis that a low-degree generator, which is more likely for larger $s$, makes the problem easier, but this is eventually counteracted by the fact that increasing $s$ always increases the minimum number of polynomial additions required. The fact that {\tt weighted} is easier than {\tt uniform} across values of $d$ and $s$ also supports this hypothesis.


\textbf{Difficulty increases relatively slowly with $d$:} The growth appears to be either linear or slightly sub-linear in $d$ in the range tested (Figures \ref{fig:3-weighted-additions} and \ref{fig:3-uniform-additions}).

\textbf{Zero dimensional ideals are rare:} (Table \ref{tab:dimension}) For $n=3$, $d=20$, the hardest distribution is $s=4$, in which case $.05\%$ of the ideals were zero dimensional. This increased to $21.2\%$ using the {\tt weighted} distribution and increasing to $s=10$, which is still relatively rare. This also supports the hypothesis that the appearance of a generator of low degree makes the problem easier.

\begin{table}[h]
  \caption{Dimension of the binomial ideals (i.e., the dimension of the solution set of the corresponding system of equations), in a sample of 10000 ($n=3$, $d=20$).}
  \label{tab:dimension}
  \vskip 0.15in
  \begin{center}
  \begin{small}
  \begin{sc}
  \begingroup
  \setlength{\tabcolsep}{5pt}
  \begin{tabular}{r|cccc}
  \toprule
  & \multicolumn{2}{c}{weighted} & \multicolumn{2}{c}{uniform} \\
  $\dim$ & $s=10$ & $s=4$ & $s=10$ & $s=4$ \\
  \midrule
  0 & 2121 & 178 & 58 & 5 \\
  1 & 7657 & 6231 & 8146 & 2932 \\
  2 & 223 & 3592 & 1797 & 7064 \\
  \bottomrule
  \end{tabular}
  \endgroup
  \end{sc}
  \end{small}
  \end{center}
  \vskip -0.1in
\end{table}

\section{Experimental Setup}\label{sec:experimentalsetup}

We train a neural network model to perform pair selection in Buchberger's algorithm.



\subsection{Network Structure}


We represent a state $S_t = (G_t, P_t)$ as a matrix whose rows are obtained by concatenating the exponent vector of each pair. For $n$ variables and $p$ pairs, this results in a matrix of size $p \times 4n$. The environment is now partially observed, as the observation does not include the coefficients.


\begin{ex}
Let $n = 3$, and consider the state given by $G = \{xy^6 + 9y^2z^4, z^4 + 13z, xy^3 + 91xy^2\}$, where the terms of each binomial are shown in grevlex order, and $P = \{ (1, 2), (1, 3), (2, 3) \}$. Mapping each pair to a row yields
\[
\left[\begin{array}{@{}cccccc|cccccc@{}}
1 & 6 & 0 & 0 & 2 & 4 & 0 & 0 & 4 & 0 & 0 & 1 \\
1 & 6 & 0 & 0 & 2 & 4 & 1 & 3 & 0 & 1 & 2 & 0 \\
0 & 0 & 4 & 0 & 0 & 1 & 1 & 3 & 0 & 1 & 2 & 0 \\
\end{array}\right]
\]
\end{ex}

Our agent uses a policy network that maps each row to a single preference score using a series of dense layers. We implement these layers as 1D convolutions with $1\times 1$ kernel in order to compute the preference score for all pairs simultaneously. The agent's policy, which is a probability distribution on the current pair set, is the softmax of these preference scores. In preliminary experiments, network depth did not appear to significantly affect performance, so we settled on the following architecture:
\[
\xymatrix{
*+[F-:<3pt>]{p \times 4n} \ar[r]^-*++{\txt{1D conv \\ relu}} & *+[F-:<3pt>]{p \times 128} \ar[r]^-*++{\txt{1D conv \\ linear}} & *+[F-:<3pt>]{p\times 1} \ar[r]^-*++{\text{softmax}} & *+[F-:<3pt>]{p \times 1}
}
\]

Due to its simplicity, it would in principle be easy to deploy this model in a production implementation of Buchberger's algorithm. The preference scores produced by the network could be used as sort keys for the pair set. Each pair would only need to be processed once, and we expect the relatively small matrix multiplies in this model to add minimal overhead in a careful implementation. In fact, most of the improvement was already achieved by a model with only four hidden units (see supplement).

However, given that real time performance of Buchberger's algorithm is highly dependent on sophisticated implementation details, we exclusively focus on implementation independent metrics, and defer the testing of real time performance improvements to future work.

\begin{table*}[h!]
  \caption{Agent performance versus benchmark strategies in 3 variables and degree 20. Each line is a unique agent trained on the given distribution. Performance is mean[stddev] on 10000 new randomly sampled ideals from that distribution. Training times were 16 to 48 hours each on a c5n.xlarge instance through Amazon Web Services. Smaller numbers are better.}
  \label{tab:performance}
  \vskip 0.15in
  \begin{center}
  \begin{small}
  \begin{sc}
  \begin{tabular}{cc|ccccc|cc}
    \toprule
    $s$ & distribution & First & Degree & Normal & Sugar & Random & Agent & Improvement \\
    \midrule
    10 & weighted & 187.[73.1] & 136.[50.9] & 136.[51.2] & 161.[66.9] & 178.[68.3] & 85.6[27.3] & 37\% [46\%] \\
    4 & weighted & 210.[101.] & 160.[64.5] & 160.[66.6] & 185.[87.2] & 203.[97.8] & 101.[44.9] & 37\% [30\%] \\
    10 & uniform & 352.[117.] & 197.[55.7] & 198.[57.1] & 264.[88.5] & 318.[103.] & 141.[42.8] & 28\% [23\%] \\
    4 & uniform & 317.[130.] & 195.[70.0] & 194.[70.0] & 265.[107.] & 303.[122.] & 151.[56.4] & 22\% [19\%] \\
    \bottomrule
  \end{tabular}
  \end{sc}
  \end{small}
  \end{center}
  \vskip -0.1in
\end{table*}


\subsection{Value Functions}

\begin{figure}[h]
  \vskip 0.2in
  \begin{center}
  \centerline{\includegraphics[width=0.95\columnwidth]{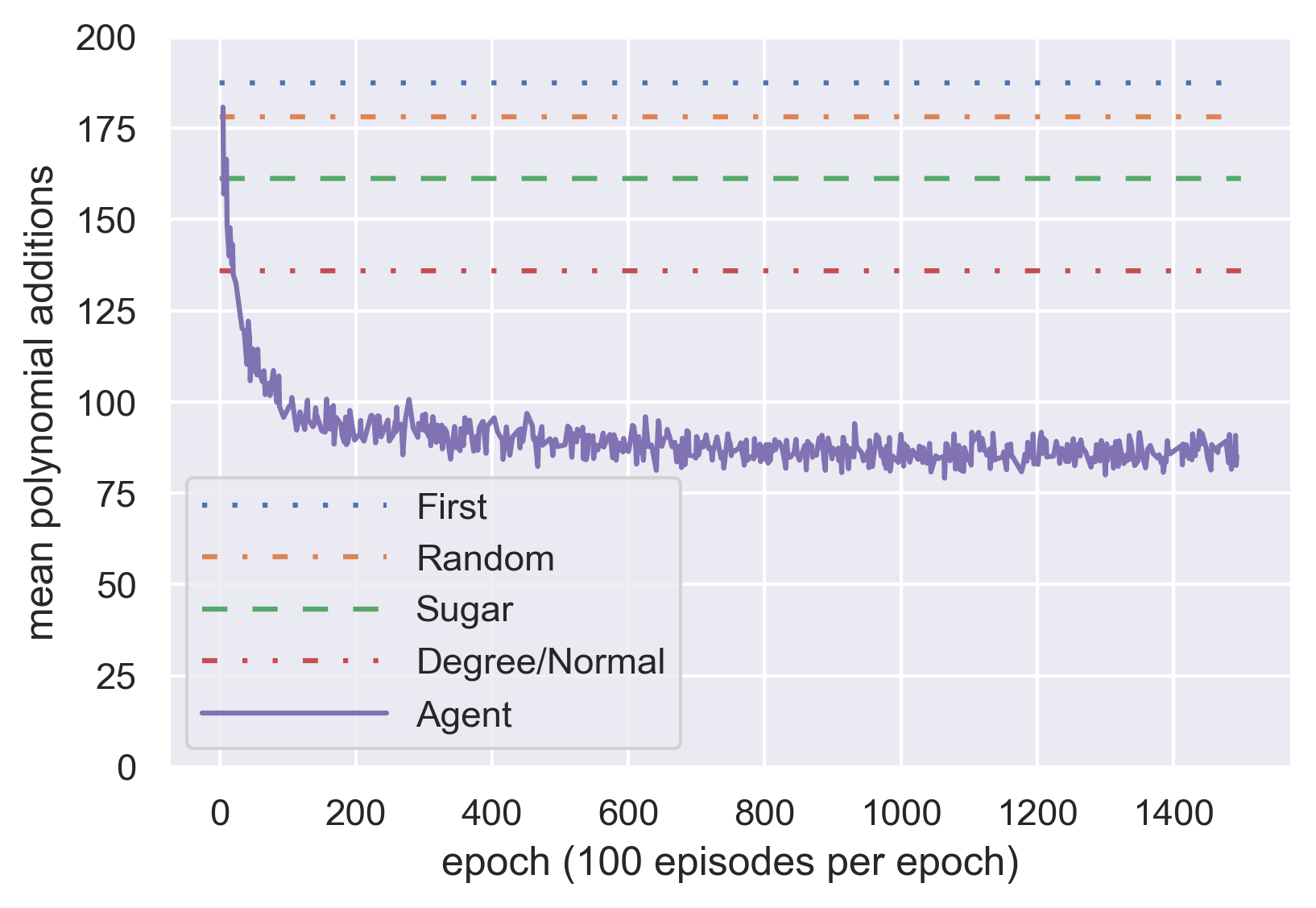}}
  \caption{Mean performance during each epoch of training on the 3-20-10 weighted distribution. Dashed lines indicate mean performance of benchmark strategies on 10000 random ideals. Total training time was 16 hours on a c5n.xlarge instance through Amazon Web Services. Smaller numbers are better.}
  \label{fig:training}
  \end{center}
  \vskip -0.2in
\end{figure}

A general challenge for policy gradient algorithms is the large variance in the estimate of expected rewards. This is exacerbated in our context by the large variance in difficulty of computing a Gr\"{o}bner basis of different ideals from the same distribution. We address this using Generalized Advantage Estimation (GAE) \cite{SchulmanEtAl2016}, which uses a value function to produce a lower-variance estimator of expected returns while limiting the bias introduced. Our value function $V(G,P)$ is the number of polynomial additions required to complete a full run of Buchberger's algorithm starting with the state $(G,P)$, using the Degree strategy. This is computationally expensive but significantly improves performance.

\subsection{Training Algorithm}

Our agents are trained with proximal policy optimization (PPO) \cite{SchulmanEtAl2017} using a custom implementation inspired by \cite{SpinningUp}.
In each epoch we first sample 100 episodes following the current policy.
 Next, GAE with $\lambda = 0.97$ and $\gamma = 0.99$ is used to compute advantages, which are normalized over the epoch.
Finally, we perform at most 80 gradient updates on the clipped surrogate PPO objective with $\epsilon = 0.2$ using Adam optimization with learning rate $0.0001$.
Early-stopping is performed when the sampled KL-divergence from the last policy exceeds 0.01.

\begin{figure}[ht]
  \vskip 0.22in
  \begin{center}
  \centerline{\includegraphics[width=\columnwidth]{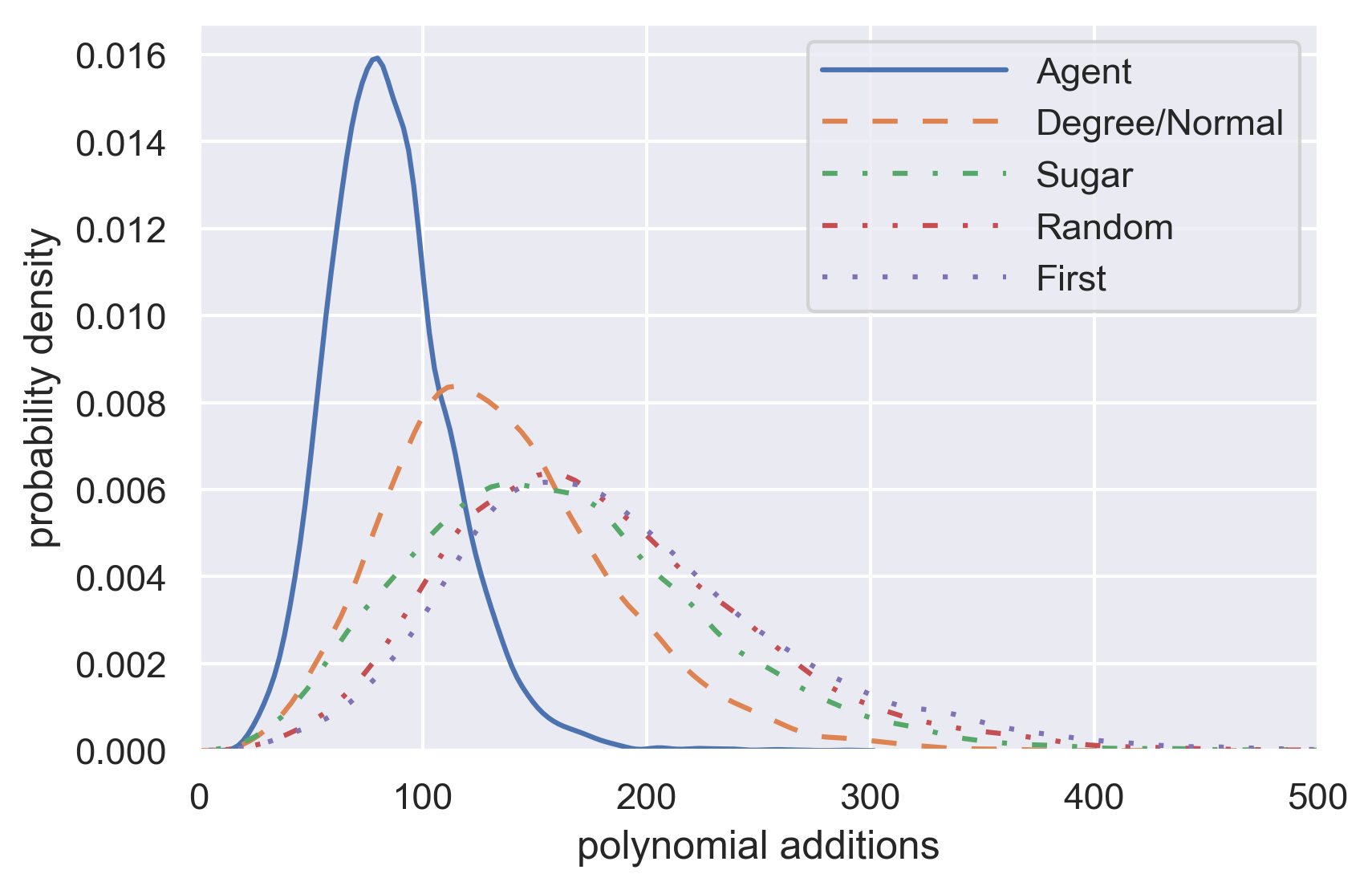}}
  \caption{Estimated distribution of polynomial additions per ideal in the 3-20-10 weighted distribution for the fully trained agent from Figure~\ref{fig:training}, compared to benchmark strategies. Smaller numbers are better. (10000 samples, computed using kernel density estimation)}
  \label{fig:trained-dist}
  \end{center}
  \vskip -0.2in
\end{figure}

\subsection{Data Generation}

There are no fixed train or test sets. Instead, the training and testing data are generated online by a function that builds an ideal at random from the distribution. The large size of the distributions prevents any over-fitting to a particular subset of ideals. For example, even ignoring coefficients, the total number of ideals in 3-20-10 weighted is roughly $10^{55}$. The agent trained in Figure~\ref{fig:training} saw 150000 ideals from this set generated at random during training. This agent was tested by running on a completely new generated set of 10000 ideals to produce the results in Table~\ref{tab:performance}.

\section{Experimental Results}\label{sec:experimentalresults}

Table \ref{tab:performance} shows the final performance of agents which have been trained on several distributions with $n=3$, $d=20$. All agents use 22\% to 37\% fewer polynomial additions on average than the best benchmark strategies, and reduce the standard deviation in the number of polynomial additions by 19\% to 46\%. The improvement on {\tt uniform} distributions, which tend to produce ideals of higher average difficulty, is not as large as the improvement on {\tt weighted} distributions. Figure~\ref{fig:trained-dist} gives a more detailed view of the distribution of polynomial additions per ideal performed by the trained agent. Figure~\ref{fig:training} shows the rapid convergence during training.

\subsection{Interpretation}\label{subsec:interpretation}

We have identified several components of the agents strategy: (a) the agent is mimicking Degree, (b) the agent prefers pairs whose $S$-polynomials are monomials, (c) the agent prefers pairs whose $S$-polynomials are low degree.

On 10000 sample runs of Buchberger's algorithm using a trained agent on 3-20-10 weighted, the average probability that the agent selected a pair which could be chosen by Degree was 43.5\%. If there was a pair in the list whose $S$-polynomial was a monomial, the agent picked such a pair 31.7\% of the time. The probability that the agent selected a pair whose $S$-polynomial had minimal degree (among $S$-polynomials), was 48.3\%.

It is notable that (b) and (c) are not standard selection heuristics. When we hard-coded the strategy of selecting a pair with minimal degree $S$-polynomial, which we call TrueDegree, the average number of additions (3-20-10 weighted, 10000 samples) was 120.3, a 12\% improvement over the Degree strategy. On the other hand, for the strategy which follows Degree but will first select any $S$-polynomial which is monomial, the average number of additions was 134.2, a 1.2\% improvement over Degree.
While neither hard-coded strategy achieves the 37\% improvement of the agent over Degree, it is notable that these insights from the model led to understandable strategies that beat our benchmark strategies in this domain.


\subsection{Variants of the Model}

We found that the model performance decreased when we made any of the following modifications: only allowed the network to see the lead monomials of each pair; removed the value function; or substituted the value function with a naive ``pairs left" value function which assigned $V(G,P) = |P|$. See Table \ref{tab:variants}. However, all of these trained models still outperform the best benchmark strategy, which is Degree.


\begin{figure}[h]
  \vskip 0.2in
  \begin{center}
  \centerline{\includegraphics[width=\columnwidth]{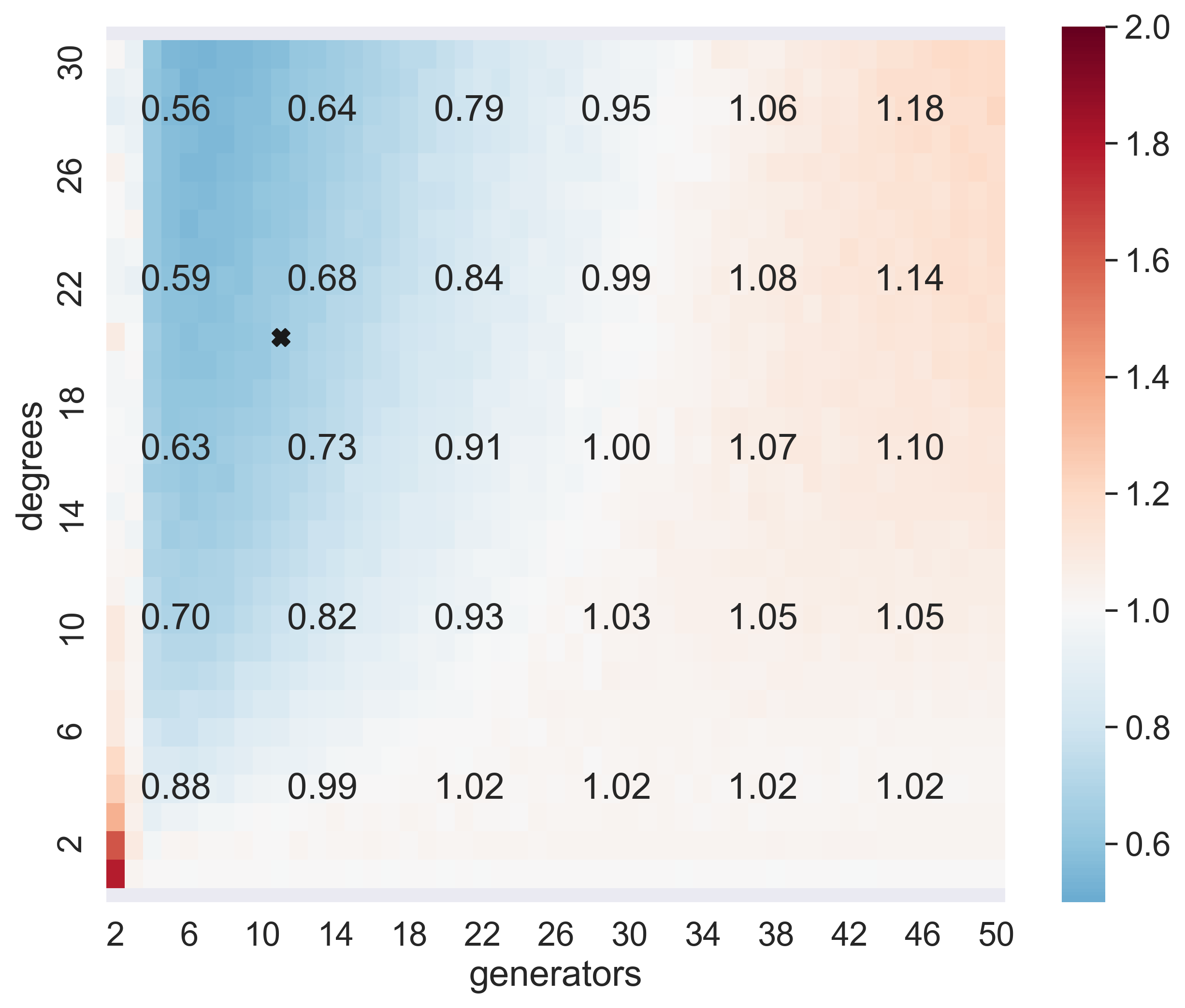}}
  \caption{Testing a single agent on 3-$d$-$s$ weighted distribution as $d$ and $s$ vary. Agent is trained on 3-20-10 weighted, indicated with an ``X." Numbers are the ratio of mean polynomial additions by the agent to that of the best benchmark strategy, with numbers less than 1 indicating better performance by the agent. The agent was tested on 1000 random ideals in each distribution, and the strategies were tested on 10000 for $s \le 20$ and 1000 for $s > 20$.}
  \label{fig:generalization-weighted}
  \end{center}
  \vskip -0.2in
\end{figure}

\begin{table}[h]
  \caption{Performance of variants of the model. Entries show mean [stddev] of polynomial additions and performance drop relative to the original model on samples of 10000 ideals from 3-20-10 weighted distribution. Original model is 85.6 [27.3].}
  \label{tab:variants}
  \vskip 0.15in
  \begin{center}
  \begin{small}
  \begin{sc}
  \begin{tabular}{r|cc}
    \toprule
    Agent & Additions & Drop \\
    \midrule
    pairsleft value function & 95.2 [32.7] & 11.2\% \\
    no value function & 103.2 [35.9] & 20.6\% \\
    \hline
    lead monomial only & 90.0[29.4] & 5.4\% \\
    \bottomrule
  \end{tabular}
  \end{sc}
  \end{small}
  \end{center}
  \vskip -0.1in
\end{table}

\begin{table}[h!]
  \caption{Agent performance outside of training distribution. Performance is mean[stddev] on a sample of 10000 random ideals.}
  \label{tab:generalization}
  \vskip 0.15in
  \begin{center}
  \begin{small}
  \begin{sc}
  3-20-10 distribution\\
  \begin{tabular}{r|cc}
   \toprule
   \backslashbox{train}{test} & weighted & uniform \\
   \midrule
   weighted & 85.6[27.3] & 140.[45.7] \\
   uniform & 89.3[29.0] & 141.[42.8] \\
   \bottomrule
  \end{tabular}
  \vskip 0.15in
  3-20-4 distribution\\
  \begin{tabular}{r|cc}
   \toprule
   \backslashbox{train}{test} & weighted & uniform \\
   \midrule
   weighted & 101.[44.9] & 158.[67.9] \\
   uniform & 107.[42.6] & 151.[56.4]  \\
   \bottomrule
   \end{tabular}
  \end{sc}
  \end{small}
  \end{center}
  \vskip -0.1in
\end{table}

\subsection{Generalization across Distributions}

A major question in machine learning is the ability of a model to generalize outside of its training distribution. Table~\ref{tab:generalization} shows reasonable generalization between {\tt uniform} and {\tt weighted} distributions.

Figure~\ref{fig:generalization-weighted} shows that a model trained on 3-20-10 weighted has similar performance at nearby values of $d$ and $s$, as compared to the performance of the best benchmark strategy.
Agents can also be trained on a mix of distributions by randomly selecting a training distribution at each epoch.
Choosing uniformly from $5 \le d \le 30$ and $4 \le s \le 20$ yields agents with 1-5\% worse performance at 3-20-10 weighted and 1-10\% better performance away from it, though performance does eventually degrade as in Figure~\ref{fig:generalization-weighted}.

\subsection{Future directions}

It would be interesting to extend these results to more variables and non-binomial ideals. In the interest of establishing a simple proof-of-concept, we have left a thorough investigation of these questions for future research, but we have done some preliminary experiments.


In the direction of increasing $n$, we trained and tested our model (with the same hyperparameters) on binomial ideals in five variables. The agents use on average 48\% fewer polynomial additions than the best benchmark strategy in the 5-10-10 weighted distribution, 28\% fewer in 5-5-10 weighted, and 11\% fewer in the 5-5-10 uniform distribution. We could not increase the degree further or perform a full hyperparameter search due to computational constraints.


In the non-binomial setting, we tested our agent on a toy model for sparse polynomials. We sampled generators for our random ideals by drawing a binomial from the {\tt weighted} distribution, then adding $k$ monomial terms drawn from the same distribution, where $k$ is sampled from a Poisson distribution with parameter $\lambda$.

\begin{figure}[ht!]
  \vskip 0.2in
  \begin{center}
  \centerline{\includegraphics[width=\columnwidth]{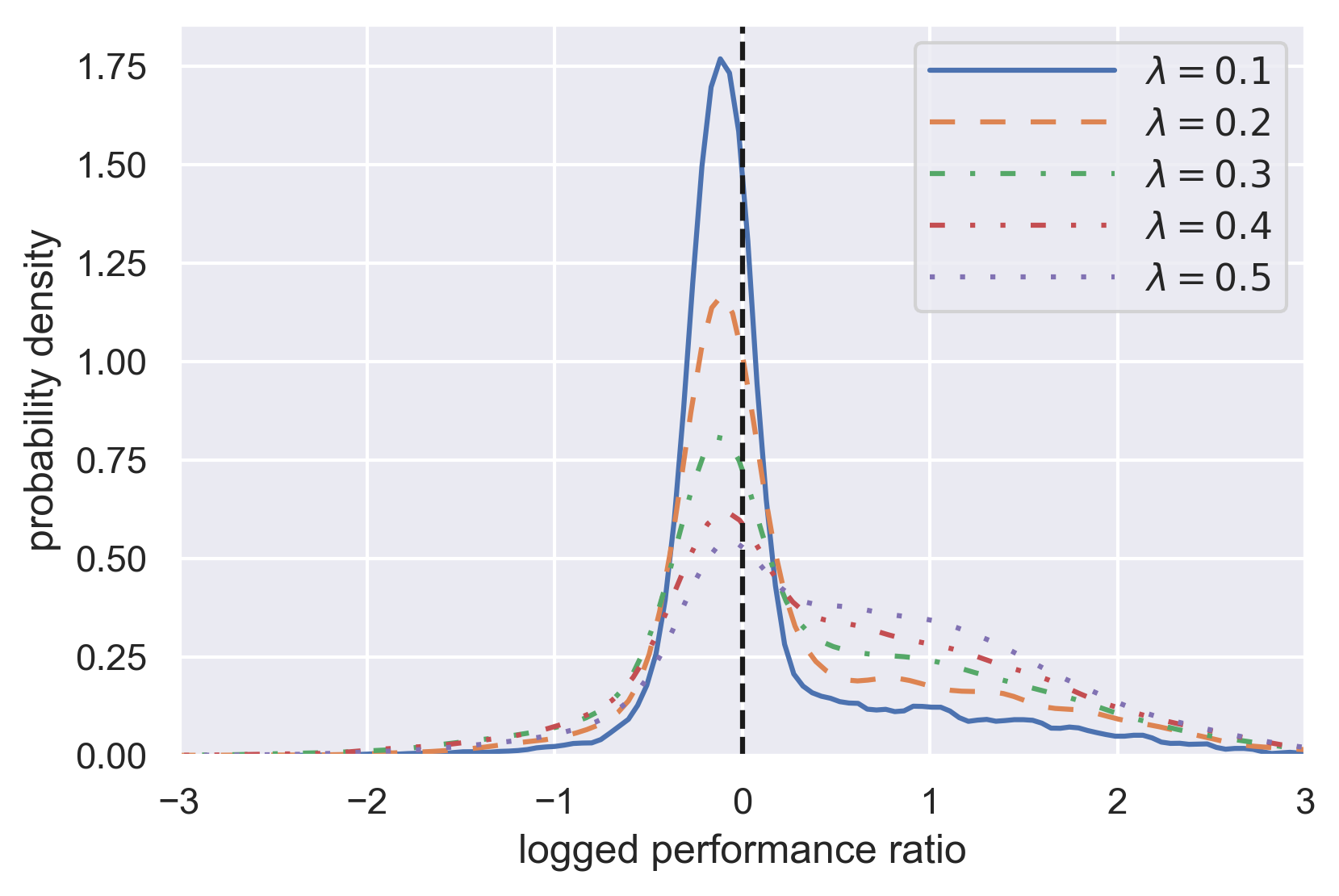}}
  \caption{Agent performance on non-binomial ideals. The logged performance ratio is the base-10 log of agent polynomial additions to best benchmark strategy on each of a sample of 10000 ideals. Values less than $0$ indicate better performance by the agent.}
  \label{fig:generalization-polynomials1}
   \end{center}
  \vskip -0.2in
\end{figure}
\begin{figure}[ht!]
  \vskip 0.2in
  \begin{center}
  \centerline{\includegraphics[width=\columnwidth]{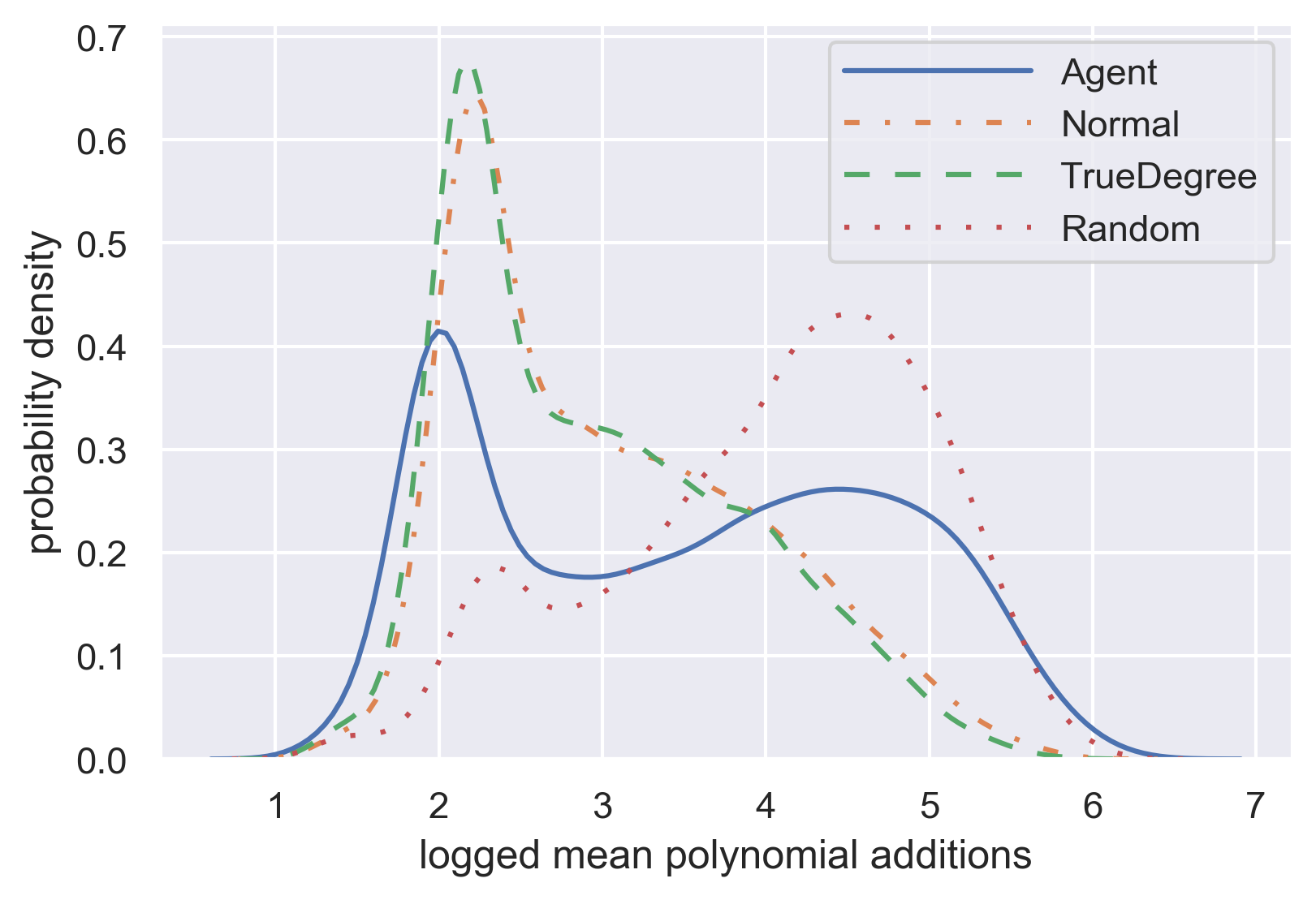}}
  \caption{Estimated distribution of base-10 log of polynomial additions per ideal with $\lambda = 0.5$, compared to benchmark strategies. (10000 samples, computed using kernel density estimation)}
  \label{fig:generalization-polynomials2}
  \end{center}
  \vskip -0.2in
\end{figure}

The fully trained agent from Figure~\ref{fig:training} had mixed results when tested on this non-binomial distribution. The distribution for the agent's performance is bimodal, with it outperforming all benchmarks on many ideals but behaving essentially randomly on others, see Figure~\ref{fig:generalization-polynomials1} and Figure~\ref{fig:generalization-polynomials2}. As a result, the agent significantly underperformed the best benchmark on average, see Table~\ref{tab:nonbinom}, but still had the best median performance for $\lambda=0.1,0.2$. TrueDegree, the strategy derived from the model in Section~\ref{subsec:interpretation}, outperforms the best benchmark in mean by 19\% to 33\% for all $\lambda$.

\begin{table}[h]
  \caption{Mean polynomial additions of several strategies tested on samples of 10000 non-binomial ideals. Agent was trained on the 3-20-10 weighted binomial distribution. Benchmark refers to the Normal strategy, the best performing benchmark in this case.}
  \label{tab:nonbinom}
  \vskip 0.15in
  \begin{center}
  \begin{small}
  \begin{sc}
  \begin{tabular}{r|ccc}
    \toprule
    $\lambda$ & Agent & TrueDegree & Benchmark \\
    \midrule
    0.1 & 4.17e+3 & 625.  & 872. \\
    0.3 & 2.16e+4 & 3138. & 4693. \\
    0.5 & 4.96e+4 & 8436. & 1.14e+4 \\
    \bottomrule
  \end{tabular}
  \end{sc}
  \end{small}
  \end{center}
  \vskip -0.1in
\end{table}


Finally, the value function used for training with GAE, one of the main contributors to our performance improvement, effectively squares the complexity by doing a full rollout at every step. Therefore, a more efficient modeled value function is crucial for scaling these results both to higher numbers of variables and to non-binomials.



\section{Conclusion}

We have introduced the Buchberger environment, a challenging reinforcement learning problem with important ramifications for the performance of computer algebra software. We have identified binomial ideals as an interesting domain for this problem that is tractable, maintains many of the problem's interesting features, and can serve as a benchmark for future research.

Standard reinforcement learning algorithms with simple models can develop strategies that improve over state-of-the-art in this domain. This illustrates a direction in which modern developments in machine learning can improve the performance of critical algorithms in symbolic computation.

\section*{Acknowledgements}

We thank the anonymous reviewers for their helpful feedback and corrections, and David Eisenbud for useful discussions. Dylan Peifer and Michael Stillman were partially supported by NSF Grant No.~DMS-1502294, and Daniel Halpern-Leistner was partially supported by NSF Grant No.~DMS-1762669.

\bibliography{PeiferEtAl2020}
\bibliographystyle{icml2020}

\newpage
\section*{Supplementary Material}
\renewcommand{\thesection}{\Alph{section}}
\setcounter{section}{0}

\section{Experimental methods}

The code used to generate all statistics and results for this paper is available at

{\small{\url{https://github.com/dylanpeifer/deepgroebner}}}

with selected computed statistics and training run data at

{\small{\url{https://doi.org/10.5281/zenodo.3676044}}}

Statistics for First, Degree, Normal, Sugar, and Random were generated using Macaulay2 1.14 on Ubuntu 18.04 while agent training was performed on c5n.xlarge instances from Amazon Web Services using the Ubuntu 18.04 Deep Learning AMI.

There were four primary training settings corresponding to the four distributions in Table~3 from Section~5.
In each setting we performed three complete runs using the parameters from Table~\ref{tab:parameters} below.
Model weights were saved every 100 epochs, and a single model was selected from the available runs and save points in each setting based on best smoothed training performance.

\begin{table}[h]
  \caption{Hyperparameters for primary evaluation runs.}
  \label{tab:parameters}
  \vskip 0.15in
  \begin{center}
  \begin{small}
  \begin{sc}
  \begin{tabular}{r|c}
    \toprule
    hyperparameter & value \\
    \midrule
    $\gamma$ for GAE & 0.99 \\
    $\lambda$ for GAE & 0.97 \\
    $\epsilon$ for PPO & 0.2 \\
    optimizer & Adam \\
    learning rate & 0.0001 \\
    max policy updates per epoch & 80 \\
    policy KL-divergence limit & 0.01 \\
    hidden layers & [128] \\
    value function & degree agent \\
    epochs & 2500 \\
    episodes per epoch & 100 \\
    max episode length & 500 \\
    \bottomrule
  \end{tabular}
  \end{sc}
  \end{small}
  \end{center}
  \vskip -0.1in
\end{table}

Trained models were then evaluated on new sets of ideals to produce Table~3 in Section~5, Table~5 in Section~5.3, and Figure~6 in Section~5.3.
The three models from Table~4 in Section~5.2 were selected and evaluated in the same way, but were trained with their respective modifications.

\section{Hyperparameter tuning}

In addition to the main evaluation runs for this paper, we performed a brief hyperparameter search in two stages.
Both stages were trained in the 3-20-10 weighted distribution as it gives the fastest training runs.

In the first stage we varied the parameters $\gamma$ in  $\{1.0, 0.99\}$, $\lambda$ in $\{1.0, 0.97\}$, learning rate in $\{10^{-3}, 10^{-4}, 10^{-5}\}$, and network as a single hidden layer of 128 units or two hidden layers of 64 units.
Three runs were performed on each set of parameters, for a total of 72 runs.
The pairs left value function was used in this search instead of the degree agent, as it leads to significantly faster training.
Learning rates of $10^{-4}$ showed best performance, though rates of $10^{-5}$ were still improving at the end of the runs.
Changes in $\gamma$, $\lambda$, and network did not consistently change performance.

In the second stage we varied just the network shape.
Single hidden layer networks were tried with 4, 8, \dots, 256 units and two hidden layer networks were tried with 4, 8, \dots, 256 hidden units in each layer.
One run was performed on each network, for a total of 14 runs.
Results showed significant improvement in using at least 32 hidden units and no major differences between one and two hidden layers. Small models were also surprisingly effective, with the model with a single hidden layer of 4 units achieving mean performance of around 100 polynomial additions during training, compared to Degree selection at 136 and our full model at 85.6.

\section{Complexity analysis of Buchberger's algorithm}

In this section, we consider upper and lower bounds for the complexity of 
computing a Gr\"obner basis of an ideal in a polynomial ring, and also describe what happens in the "generic" (random) case, giving more detail than in the paper proper.  We first consider the maximum degree of a Gr\"obner basis element, then describe how that gives bounds for the size of a minimal reduced Gr\"obner basis.

Let $I = \langle f_1, \ldots, f_s\rangle \subset S = k[x_1, \ldots, x_n]$, be an ideal, with each polynomial $f_i$ of degree $\le d$.  If $>$ is a monomial order, and $G = GB_>(I) = \{ g_1, \ldots, g_r \}$ is a Gr\"obner basis of $I$, $G$ is called a \emph{minimal and reduced} Gr\"obner basis if the lead coefficient of each $g_i$ is one, and no monomial of $g_i$ is divisible by $LT(g_j)$, for $i \ne j$.  Given any Gr\"obner basis, it is easy to modify $G$ to obtain a minimal and reduced Gr\"obner basis of $I$.  Given the monomial order, each ideal $I$ has precisely one minimal reduced Gr\"obner basis with respect to this order.  We define $\deg_{\max}(GB_>(I)) := \max(\deg g_1, \ldots, \deg g_r)$, where $G = GB_>(I) = \{ g_1, \ldots, g_r\}$ is the unique minimal reduced Gr\"obner basis of $I$ for the order $>$.

\subsection*{Upper bounds}
We have the following upper bound for $\deg_{\max}(GB_>(I))$.
\begin{thm}[\cite{Dube1990}]
Given $I$ as above, then 
\[ \deg_{\max}(GB_>(I)) \,\le\, 2\, (\frac{d^2}{2} + d)^{2^{n-1}}\]
If $I$ is homogeneous (i.e. each polynomial $f_i$ is homogeneous), we may replace the $n-1$ by $n-2$ in this bound.
\end{thm}

Such a bound is called double exponential (in the number of variables).  This result seems to give an incredibly bad bound, but it is unfortunately fairly tight, which we will discuss next.
\subsection*{Lower bounds}

All known double exponential examples (e.g. \cite{BayerStillman88}, \cite{Koh98}, \cite{Mora2005}) are based essentially on the seminal and important construction of \cite{MayrMeyer82}.  Each is a sequence of ideals $J_n$ where the $n$-th ideal $J_n$ is in roughly $10n$ or $20n$ variables, generated in degrees $\le d$, where each $f_i$ is a \emph{pure binomial} (i.e. a difference of two monomials).  The following version of \cite{Koh98} is a sequence of ideals
generated by \emph{quadratic} pure binomials.

\begin{thm}[\cite{Koh98}]
For each $n \ge 1$, there exists an ideal $J_n$, generated by quadratic homogeneous pure binomials in $22n-1$ variables such that for any monomial order $>$,
\[ 2^{2^{n-1}-1} \le  \deg_{\max}(GB_>(J_n)) \]
\end{thm}
\cite{Koh98} shows that there is a minimal syzygy in degree $2^{2^{n-1}}$.  It is well known (see e.g. \cite{Mora2005}, section 38.1) that this implies that there must be a minimal Gr\"obner basis element of degree at least half that, giving the stated bound.  Thus there are examples of ideals whose Gr\"obner basis has maximum degree bounded below by roughly $2^{2^{n/22}}$ (where now $n$ is the number of variables).  Some of the other modifications of the Mayr-Meyer construction have slightly higher lower bounds (e.g.$ 2^{2^{n/10}}$).

\subsection*{Better bounds}
Given these very large lower bounds, one might conclude that Gr\"obner bases cannot be used in practice.  However, in many cases, there exist much lower upper bounds for the size of a grevlex Gr\"obner basis.  The key is to relate these degree bounds to the \emph{regularity} of the 
ideal.

Given a homogeneous ideal $I = \langle f_1, \ldots, f_s\rangle \subset S = k[x_1, \ldots, x_n]$, with each polynomial of degree $\le d$, several notions which often appear in complexity bounds and are also useful in algebraic geometry are: 
\begin{itemize}
\item the dimension $dim(I)$ of $I$. 
\item the depth $depth(I)$.  This is an integer in the range $0 \le depth(I) \le dim(I)$.  In many commutative algebra texts, this is denoted as $depth(S/I)$, not $depth(I)$, but in \cite{Mora2005}, $depth(I)$ is the notation.  This is the length of a maximal $S/I$-regular sequence in $(x_1, \ldots, x_n)$.
\item the (Castelnuovo-Mumford) regularity, $reg(I)$ of the ideal $I$, see \cite{Eisenbud1995} or \cite{Mora2005}.
\end{itemize}

The regularity $reg(I)$ should be considered as a measure of complexity of the ideal.

\subsubsection*{Generic change of coordinates}
Let's consider a homogeneous, linear, change of coordinates $\phi = \phi_A$, where $A \in k^{n \times n}$ is a square $n$ by $n$ matrix over $k$, with
\[ \phi_A(x_i) = \sum_{j=1}^{n} A_{ij} x_j. \]
Let $\phi_A(I) := \{ f(\phi_A(x_1), \ldots, \phi_A(x_n)) \mid f \in I \}$
be the ideal under a change of coordinates.  Consider the $n^2$-dimensional parameter space $V$ (where a point $A = (A_{ij})$ of $V$ corresponds to a homogeneous linear change of coordinates $\phi_A$).  It turns out that there is a polynomial $F$ in the polynomial ring (with $n^2$ variables) $k[A_{ij}]$, such that for all points $A \in V$ such that $F(A) \ne 0$, then $LT_{grevlex}(\phi_A(I))$ is the same ideal.  This monomial ideal is called the \emph{generic initial ideal} of $I$ (in grevlex coordinates), and is denoted by $gin(I)$.  Basically, for a random homogeneous linear change of coordinates, one always gets the same size Gr\"obner basis, with the same lead monomials.

Define $G(I)$ to be the maximum degree of a minimal generator of $gin(I)$.  This is the maximum degree of an element of the unique minimal and reduced Gr\"obner basis of the ideal $\phi_A(I)$ under almost all change of coordinates $\phi_A$ (i.e. those for which $F(A) \ne 0$).

The reason this is important is that we have more control over Gr\"obner bases in generic coordinates.  For instance

\begin{thm}[\cite{BayerStillman1987}]
If the base field is infinite, then
\[ reg(I) = reg(gin(I))
\]
If the characteristic of $k$ is zero, then 
\[ G(I) = reg(I)
\]
If the characteristic of $k$ is positive, then 
\[ \frac{1}{n} reg(I) \le G(I) \le reg(I)
\]
\end{thm}

It is know that $reg(I) \le reg(LT_>(I))$, for every monomial order $>$.  This result states that in fact in generic coordinates, equality is obtained for the grevlex order.  The Gr\"obner basis, after a random change of coordinates, always has maximum degree at most $reg(I)$.

In particular, if an ideal $I$ has small regularity, as often happens for ideals coming from algebraic geometric problems, then the corresponding Gr\"obner basis in grevlex order will have much smaller size than the double exponential upper bounds suggest.

\begin{thm}
If the homogeneous ideal $I = \langle f_1, \ldots, f_s\rangle \subset S$ has $\dim(I) = depth(I)$ (this includes the case when $dim(I) = 0$, then
\[ reg(I) \le (d-1) \min(s,n - \dim(I))) + 1                       
\]
\end{thm}

This follows from two basic facts about regularity: First, if $\dim I = 0$, then the regularity of $I$ is the first degree $m$ such that the degree $m$ polynomials in $I$ consist of all degree $m$ polynomials.  Second, if $I$ has depth $r$ and $y_1, \ldots, y_r $ is a regular sequence of linear forms mod $I$, then the regularity of $I$ is the regularity of the ideal $\overline{I} := I S/(y_1, \ldots, y_r)$.  Since the depth and dimension of $I$ are equal, the ideal $\overline{I}$ is of dimension $0$, and contains a complete intersection of polynomials each of degree $d$.  This implies by a Hilbert function argument, or by the Koszul complex, that the regularity of $\overline{I}$ is at most $(d-1)(n-r) + 1$ (see \cite{Eisenbud1995} for these kinds of arguments).

This implies that $G(I) \le  (d-1) \min(s,n-\dim(I)) + 1 \le dn$, a dramatic improvement on the double exponential bounds!

\subsection*{Ideals generated by random, or generic, polynomials}
What happens for random homogeneous ideals generated by $s$ polynomials each of degree $d$? For fixed $n,d,s$, the space of possible inputs, i.e., the space $V$ of coefficients for each of the $s$ generators, is finite dimensional. There is a subset $X \subset V$, a closed algebraic set (so having measure zero, if the base field is $\mathbb{R}$ or $\mathbb{C}$), such that for any point outside $X$, the corresponding ideal $I$ satisfies
$\dim(I) = depth(I)$, and therefore,
\[ G(I) \le (d-1) \min(s,n) + 1. \]
In characteristic zero, equality holds.

If instead of homogeneous ideals, we consider random inhomogeneous ideals,  generated by $s$ polynomials each of degree $d$.  The same method holds: the homogenization of these polynomials puts us into the situation in the previous paragraph.  Therefore for such inhomogeneous ideals, whose coefficient point is outside of $X$, then the ideal $J$ generated by the homogenization of the $f_i$ with respect to a new variable satisfies 
$\dim(J) = depth(J)$, and therefore,
\[ G(I) = G(J) \le (d-1) \min(s,n+1) + 1. \]
In characteristic zero, equality holds.

\subsection*{Bounds on the size of the reduced minimal Gr\"obner basis}

In the unique reduced minimla Gr\"obner basis of an ideal $I$, there can not be two generators with identical lead monomials. It follows that if all generators in this Gr\"obner basis have degree $\leq D$, then there are at most
\begin{align*}
\# \{\text{monomials of degree }\leq D\} &= \left( \begin{array}{c} D+n \\ n \end{array} \right) \\
&= \mathcal{O}\left((n+D)^{\min(n,D)}\right)
\end{align*}
generators in the Gr\"obner basis. If one combines this with the upper bound above on the maximum degree, $D=(d-1)\min(s,n+1)+1$, one finds the following upper bound on the size of the minimal reduced Gr\"obner basis of a generic ideal generated by $s$ polynomials of degree $\leq d$ in $n$ variables:
\[
\# \{\text{Gr\"obner basis generators}\} \leq \mathcal{O} \left( (n+1)^n d^n \right),
\]
where the simplification comes from approximating $\min(s,n+1) \leq n+1$, so that our bound is independent of $s$, and assuming $d \geq 2$.

\end{document}